\documentclass[10pt]{bmc_article}    

\usepackage{cite} 
\usepackage{url}  
\usepackage{ifthen}  
\usepackage{multicol}   
\usepackage[latin1]{inputenc} 
\usepackage{amsmath}
\usepackage{amssymb}
\usepackage{enumerate}
\usepackage[dvips]{graphicx}
\usepackage[dvips]{epsfig}
\usepackage{pstricks,pst-node}

\def\Tmax{T_{\textsl{max}}} 

\newboolean{publ}

\newenvironment{bmcformat}{\begin{raggedright}\baselineskip20pt\sloppy\setboolean{publ}{false}}{\end{raggedright}\baselineskip20pt\sloppy}

\begin{document}
\begin{bmcformat}


\title{A linear memory algorithm for Baum-Welch training}

\author{
        Istv\'an Mikl\'os$^{1,\S}$%
        \email{Istv\'an Mikl\'os - miklosi@ramet.elte.hu}%
      \and
        Irmtraud M. Meyer$^{2,\S,}$\correspondingauthor%
        \email{Irmtraud M. Meyer\correspondingauthor - irmtraud.meyer@cantab.net}%
      }

\address{\iid(1)MTA-ELTE Theoretical Biology and Ecology Group, P\'azm\'any P\'eter s\'et\'any 1/c
           1117 Budapest, Hungary\\
         \iid(2)European Bioinformatics Institute, Wellcome Trust Genome Campus, Cambridge CB10 1SD, UK\\
         \iid(\S)Joint first authors}
\maketitle


\begin{abstract}

\paragraph*{Background:}
Baum-Welch training is an expectation-maximisation algorithm for training the
emission and transition probabilities of hidden Markov models in a fully
automated way. It can be employed as long as a training set of annotated
sequences is known, and provides a rigorous way to derive parameter values
which are guaranteed to be at least locally optimal.  For complex hidden
Markov models such as pair hidden Markov models and very long training
sequences, even the most efficient algorithms for Baum-Welch training are
currently too memory-consuming. This has so far effectively prevented the
automatic parameter training of hidden Markov models that are currently used for
biological sequence analyses.

\paragraph*{Methods and results:}
We introduce a linear space algorithm for Baum-Welch training. For a hidden
Markov model with $M$ states, $T$ free transition and $E$ free emission
parameters, and an input sequence of length $L$, our new algorithm requires
$O(M)$ memory and $O(L M \Tmax\ (T + E))$ time for one Baum-Welch
iteration, where $\Tmax$ is the maximum number of states that any state is
connected to. The most memory efficient algorithm until now was the
checkpointing algorithm with $O(\log(L) M)$ memory and $O(\log(L) L M \Tmax\ 
)$ time requirement. Our novel algorithm thus renders the memory requirement
completely independent of the length of the training sequences. More
generally, for an n-hidden Markov model and n input sequences of length $L$,
the memory requirement of $O(\log(L) L^{n-1} M)$ is reduced to $O(L^{n-1} M)$
memory while the running time is changed from $O(\log(L) L^n M \Tmax\ + L^n (T
+ E))$ to $O(L^n M \Tmax\ (T + E))$.

An added advantage of our new algorithm is that a reduced time requirement can
be traded for an increased memory requirement and \emph{vice versa}, such that
for any $c \in \{1, \dots, (T + E)\}$, a time requirement of $L^n M
\Tmax\ c$ incurs a memory requirement of $L^{n-1} M (T + E - c)$.

\paragraph*{Conclusions:} 
For the large class of hidden Markov models used for example in gene
prediction, whose number of states does not scale with the length of the input
sequence, our novel algorithm can thus be both faster and more
memory-efficient than any of the existing algorithms.

\end{abstract}

\ifthenelse{\boolean{publ}}{\begin{multicols}{2}}{}

\section*{Background}

Hidden Markov Models (HMMs) are widely used in Bioinformatics \cite{dekm1998},
for example, in protein sequence alignment, protein family annotation
\cite{kbm+1994,eddy2001} and gene-finding \cite{md2002, md2004}.

When an HMM consisting of $M$ states is used to annotate an input sequence,
its predictions crucially depend on its set of emission probabilities
$\mathcal{E}$ and transition probabilities $\mathcal{T}$. This is for example the case
for the state path with the highest overall probability, the so-called optimal
state path or Viterbi path \cite{viterbi1967}, which is often reported as the
predicted annotation of the input sequence.

When a new HMM is designed, it is usually quite easy to define its states and
the transitions between them as these typically closely reflect the underlying
problem. However, it can be quite difficult to assign values to its emission
probabilities $\mathcal{E}$ and transition probabilities $\mathcal{T}$.  Ideally, they
should be set up such that the model's predictions would perfectly reproduce
the known annotation of a large and diverse set of input sequences.

The question is thus how to derive the best set of transition and emission
probabilities from a given training set of annotated sequences. Two main
scenarios have to be distinguished \cite{dekm1998}:

(1) If we know the optimal state paths that correspond to the known annotation
of the training sequences, the transition and emission probabilities can
simply be set to the respective count frequencies within these optimal state
paths, i.e.\ to their maximum likelihood estimators. If the training set is
small or not diverse enough, pseudo-counts have to be added to avoid
over-fitting.

(2) If we do not know the optimal state paths of the training sequences,
either because their annotation is unknown or because their annotation does
not unambiguously define a state path in the HMM, we can employ an expectation
maximisation (EM) algorithm \cite{dlr1977} such as the Baum-Welch algorithm
\cite{baum1972} to derive the emission and transition probabilities in an
iterative procedure which increases the overall log likelihood of the model in
each iteration and which is guaranteed to converge at least to a local
maximum. As in case (1), pseudo-counts or Dirichlet priors can be added to
avoid over-fitting when the training set is small or not diverse enough.

\section*{Methods and results}

\subsection*{Baum-Welch training}

The Baum-Welch algorithm defines an iterative procedure in which the emission
and transition probabilities in iteration $n+1$ are set to the number of times
each transition and emission is \emph{expected} to be used when analysing the
training sequences with the set of emission and transition probabilities
derived in the previous iteration $n$.

Let $T^n_{i,j}$ denote the transition probability for going from state $i$ to
state $j$ in iteration $n$, $E^n_i(y)$ the emission probability for emitting
letter $y$ in state $i$ in iteration $n$, $P(X)$ the probability of sequence
$X$, and $x_k$ the $k$th letter in input sequence $X$ which has length $L$. We
also define $X_k$ as the sequence of letters from the beginning of sequence
$X$ up to sequence position $k$, $(x_1, ...  x_k)$. $X^k$ is defined as the
sequence of letters from sequence position $k+1$ to the end of the sequence,
$(x_{k+1}, ... x_L)$.

For a given set of training sequences, $S$, the expectation maximisation update for
transition probability $T^n_{i,j}$, $T^{n+1}_{i,j}$, can then be written as

\begin{eqnarray}
T^{n+1}_{i,j} & = & \frac{\sum_{X \in S} t^n_{i,j}(X)/P(X)}{\sum_{j'} \sum_{X \in S} t^n_{i,j'}(X)/P(X)} \label{eq:T_update} \\
              &   & \textrm{where} 
\quad t^n_{i,j}(X) := \sum_{k = 1}^{L} f^n(X_k, i) T^n_{i,j} E^n_j(x_{k+1}) b^n(X^{k+1}, j) \nonumber
\end{eqnarray}

The superfix $n$ on the quantities on the right hand side indicates that they
are based on the transition probabilities $T^n_{i,j}$ and emission
probabilities $E^n_i(x_{k+1})$ of iteration $n$.  $f(X_k, i) := P(x_1, ...
x_k, s(x_k) = i)$ is the so-called forward probability of the sequence up to
and including sequence position $k$, requiring that sequence letter $x_k$ is
read by state $i$. It is equal to the sum of probabilities of all state paths
that finish in state $i$ at sequence position $k$. The probability of sequence
$X$, $P(X)$, is therefore equal to $f(X_L, \textsl{End})$. $b(X^k, i) :=
P(x_{k+1}, ...  x_L | s(x_k) = i)$ is the so-called backward probability of
the sequence from sequence position $k+1$ to the end, given that the letter at
sequence position $k$, $x_k$, is read by state $i$.  It is equal to the sum of
probabilities of all state paths that start in state $i$ at sequence position
$k$.

For a given set of training sequences, $S$, the expectation maximisation update for
emission probability $E^n_i(y)$, $E^{n+1}_i(y)$, is

\begin{eqnarray}
E^{n+1}_i(y) & = & \frac{\sum_{X \in S} e^n_i(y, X)/P(X)}{\sum_{y'} \sum_{X \in S} e^n_i(y', X)/P(X)} \label{eq:E_update} \\
              &   & \textrm{where} 
\quad e^n_i(y, X) := \sum_{k = 1}^{L} \delta_{x_k,y} f^n(X_k, i) b^n(X^k, i) \nonumber
\end{eqnarray}

$\delta$ is the usual delta function with $\delta_{x_k,y} = 1$ if $x_k = y$
and $\delta_{x_k,y} = 0$ if $x_k \ne y$. As before, the superfix $n$ on the
quantities on the right hand side indicates that they are calculated using the
transition probabilities $T^n_{i,j}$ and emission probabilities
$E^n_i(x_{k+1})$ of iteration $n$.

The forward and backward probabilities $f^n(X_k, i)$ and $b^n(X^k, i)$ can be
calculated using the forward and backward algorithms \cite{dekm1998} which are
introduced in the following section.

\subsubsection*{Baum-Welch training using the forward and backward algorithm}

The forward algorithm proposes a procedure for calculating the forward
probabilities $f(X_k, i)$ in an iterative way. $f(X_k, i)$ is the sum of
probabilities of all state paths that finish in state $i$ at sequence position
$k$.

The recursion starts with the initialisation

\begin{eqnarray}
f(X_0, i) & = & 
      \left\{ \begin{array}{ll} 
                 1 & i = \textsl{Start} \nonumber\\
                 0 & i \ne \textsl{Start} \nonumber
              \end{array}
      \right. \nonumber
\end{eqnarray}

where $\textsl{Start}$ is the number of the start state in the HMM. The
recursion proceeds towards higher sequence positions

\begin{equation}
f(X_{k+1}, i) = \sum_{j = 1}^M f(X_k, j) T_{j,i} E_i(x_{k+1}) \nonumber
\end{equation}

and terminates with

\begin{equation}
P(X) = P(X_L) = f(X_L, \textsl{End}) = \sum_{j = 1}^M f(X_L, j) T_{j,\textsl{End}} \nonumber
\end{equation}

where $\textsl{End}$ is the number of the end state in the HMM. The recursion
can be implemented as a dynamic programming procedure which works its way through a
two-dimensional matrix, starting at the start of the sequence in the
$\textsl{Start}$ state and finishing at the end of the sequence in the
$\textsl{End}$ state of the HMM.

The backward algorithm calculates the backward probabilities $b(X^k, i)$ in a
similar iterative way. $b(X^k, i)$ is the sum of probabilities of all state
paths that start in state $i$ at sequence position $k$. Opposed to the forward
algorithm the backward algorithm starts at the end of the sequence in the
$\textsl{End}$ state and finishes at the start of the sequence in the
$\textsl{Start}$ state of the HMM.

The backward algorithm starts with the initialisation

\begin{eqnarray}
b(X^L, i) & = & 
      \left\{ \begin{array}{ll} 
                 1 & i = \textsl{End} \nonumber\\
                 0 & i \ne \textsl{End} \nonumber
              \end{array}
      \right. \nonumber
\end{eqnarray}

and continues towards lower sequence positions with the recursion

\begin{equation}
b(X^k, i) = \sum_{j = 1}^M E_i(x_k) T_{i,j} b(X^{k+1}, j) \nonumber
\end{equation}

and terminates with

\begin{equation}
P(X) = b(X^1, \textsl{Start}) = \sum_{j = 1}^M T_{\textsl{Start},j} b(X^1, j)  \nonumber
\end{equation}

As can be seen in the recursion steps of the forward and backward algorithms
described above, the calculation of $f(X_{k+1}, i)$ requires at most $\Tmax$
previously calculated elements $f(X_{k}, j)$ for $j \in \{1,..M\}$. $\Tmax$ is
the maximum number of states that any state of the model is connected to.
Likewise, the calculation of $b(X^k, i)$ refers to at most $\Tmax$ elements
$b(X^{k+1}, j)$ for $j \in \{1,..M\}$.

In order to continue the calculation of the forward and backward values
$f(X_{k}, i)$ and $b(X_{k}, i)$ for all states $i \in \{1,..M\}$ along the
entire sequence, we thus only have to memorise $M$ elements. 

\subsubsection*{Baum-Welch training using the checkpointing algorithm}

Unit now, the checkpointing algorithm \cite{ghs1997,th1998, wh2000} was the
most efficient way to perform Baum-Welch training.

The basic idea of the checkpointing algorithm is to perform the forward and
backward algorithm by memorising the forward and backward values only in
$O(\sqrt{L})$ columns along the sequence dimension of the dynamic programming
table. The checkpointing algorithm starts with the forward algorithm,
retaining only the forward values in $O(\sqrt{L})$ columns. These columns
partition the dynamic programming table into $O(\sqrt{L})$ separate fields.
The checkpointing algorithm then invokes the backward algorithm which
memorises the backward values in a strip of length $O(\sqrt{L})$ as it moves
along the sequence. When the backward calculation reaches the boundary of one
field, the pre-calculated forward values of the neighbouring checkpointing
column are used to calculate the corresponding forward values for that field.
The forward and backward values of that field are then available at the same
time and are used to calculate the corresponding values for the EM update.

The checkpointing algorithm can be further refined by using embedded
checkpoints.  With an embedding level of $k$, the forward values in
$O(L^\frac{1}{k})$ columns of the initial calculation are memorised, thus
defining $O(L/L^\frac{1}{k}) = O(L^{\frac{k-1}{k}})$ long fields. When the
memory-sparse calculation of the backward values reaches the field in
question, the forward algorithm is invoked again to calculate the forward
values for $O(L^\frac{1}{k})$ additional columns within that field. This
procedure is iterated $k$ times within the thus emerging fields. In the end,
for each of the $O(L^\frac{1}{k})$-long k-sub-fields, the forward and backward
values are simultaneously available and are used to calculate the
corresponding values for the EM update. The time complexity of this algorithm
for one Baum-Welch iteration and a given training sequence of length $L$ is
$O(k L M \Tmax\ + L (T + E))$, since $k$ forward and $1$ backward algorithms have
to be invoked, and the memory complexity is $O(k L^\frac{1}{k} M)$.  For $k =
\log(L)$, this amounts to a time requirement of $O(\log(L) L M \Tmax\ + L (T + E))$
and a memory requirement of $O(\log(L) M)$, since $L^\frac{1}{\log(L)} = 
e$.

\subsubsection*{Baum-Welch training using the new algorithm}

It is not trivial to see that the quantities $T^{n+1}_{i,j}$ and
$E^{n+1}_i(y)$ of Equations~\ref{eq:T_update} and \ref{eq:E_update} can be
calculated in an even more memory-sparse way as both, the forward and the
corresponding backward probabilities are needed at the same time in order to
calculate the terms $f^n(X_k, i) T^n_{i,j} E^n_i(x_{k+1}) b^n(X^{k+1}, j)$ in
$t^n_{i,j}(X)$ and $\delta_{x_k,y} f^n(X_k, i) b^n(X^k, i)$ in $e^n_i(y, X)$
of Equations~\ref{eq:T_update} and \ref{eq:E_update}. A calculation of these
quantities for each sequence position using a memory-sparse implementation
(that would memorise only $M$ values at a time) both for the forward and
backward algorithm would require $L$-times more time, i.e.\ significantly more
time.  Also, an algorithm along the lines of the Hirschberg algorithm
\cite{hirshberg1975,mm1988} cannot be applied as we cannot halve the dynamic
programming table after the first recursion.

We here propose a new algorithm to calculate the quantities $T^{n+1}_{i,j}$
and $E^{n+1}_i(y)$ which are required for Baum-Welch training. Our algorithm
requires $O(M)$ memory and $O(L M \Tmax\ (T + E))$ time rather than
$O(\log(L) M)$ memory and $O(\log(L) L M \Tmax\ + L (T + E))$ time.

The trick for coming up with a memory efficient algorithm is to realise that

\begin{itemize}
\item[$\bullet$] $t^n_{i,j}(X)$ and $e^n_i(y, X)$ in
  Equations~\ref{eq:T_update} and \ref{eq:E_update} can be interpreted as
  a weighted sum of probabilities of state paths that satisfy certain
  constraints and that
\item[$\bullet$] the weight of each state path is equal to the number of
  times that the constraint is fulfilled.
\end{itemize}

For example, $t^n_{i,j}(X)$ in the numerator in Equation~\ref{eq:T_update} is
the weighted sum of probabilities of state paths for sequence $X$ that contain
at least one $i \rightarrow j$ transition, and the weight of each such state
path in the sum is the number of times this transition occurs in the state
path.

We now show how $t^n_{i,j}(X)$ in Equation~\ref{eq:T_update} can be calculated in
$O(M)$ memory and $O(L M \Tmax)$ time. As the superfix $n$ is only there to
remind us that the calculation of $t^n_{i,j}(X)$ is based on the transition and
emission probabilities of iteration $n$ and as this index does not change in
the calculation of $t^n_{i,j}$, we discard it for simplicity sake in the
following. 

Let $t_{i,j}(X_k, l)$ denote the weighted sum of probabilities of state paths
that finish in state $l$ at sequence position $k$ of sequence $X$ and that
contain at least one $i \rightarrow j$ transition, where the weight for each
state path is equal to its number of $i \rightarrow j$ transitions.

\textbf{Theorem 1:} The following algorithm calculates $t_{i,j}(X)$ in $O(M)$
memory and $O(L M \Tmax)$ time. $t_{i,j}(X)$ is the weighted sum of
probabilities of all state paths for sequence $X$ that have at least one $i
\rightarrow j$ transition, where the weight for each state path is equal to
its number of $i \rightarrow j$ transitions.

The algorithm starts with the initialisation

\begin{eqnarray}
f(X_0, m) &=& 
      \left\{ \begin{array}{ll} 
             1 & m = \textsl{Start} \nonumber \\
             0 & m \ne \textsl{Start} \nonumber
              \end{array}
      \right. \nonumber \\
t_{i,j}(X_0, m) &=& 0 \nonumber
\end{eqnarray}

and proceeds via the following recursion
\begin{eqnarray}
f(X_{k+1}, m) &=& \sum_{n = 1}^M f(X_k, n) T_{n,m} E_{m}(x_{k+1})\nonumber\\
t_{i,j}(X_{k+1}, m) & = & 
      \left\{ \begin{array}{ll} 
             \sum_{n = 1}^M t_{i,j}(X_k, n) T_{n,m} E_{m}(x_{k+1}) &  m \ne j\\
                    & \label{eq:tij} \\
                    & \\
             f(X_k, i) T_{i,m} E_m(x_{k+1}) +  & m = j\\
             \sum_{n = 1}^M t_{i,j}(X_k, n) T_{n,m} E_m(x_{k+1}) & 
              \end{array}
      \right. 
\end{eqnarray}

and finishes with

\begin{eqnarray}
P(X) = f(X_L, \textsl{End}) &=& \sum_{n = 1}^M f(X_L, n) T_{n,\textsl{End}} \label{eq:px} \\
t_{i,j}(X) = t_{i,j}(X_L, \textsl{End}) & = & 
      \left\{ \begin{array}{ll} 
             \sum_{n = 1}^M t_{i,j}(X_L, n) T_{n,\textsl{End}} & \textsl{End} \ne j\\
                    & \\
             f(X_L, i) T_{i,\textsl{End}} + & \textsl{End} = j\\
             \sum_{n = 1}^M t_{i,\textsl{End}}(X_k, n) T_{n,\textsl{End}}  & 
              \end{array}
      \right. \nonumber 
\end{eqnarray}

\textbf{Proof:} 

(1) It is obvious that the recursion requires only $O(M)$ memory as the
calculation of all $f(X_{k+1}, m)$ values with $m \in \{1,..M\}$ requires only
access to the $M$ previous $f(X_k, n)$ values with $n \in \{1,..M\}$.
Likewise, the calculations of all $t_{i,j}(X_{k+1}, m)$ values with $m \in
\{1,..M\}$ refer only to $M$ elements $t_{i,j}(X_k, n)$ with $n \in
\{1,..M\}$. We therefore have to remember only a thin ``slice'' of $t_{i,j}$
and $f$ values at sequence position $k$ in order to be able to calculate the
$t_{i,j}$ and $f$ values for the next sequence position $k+1$.  The time
requirement to calculate $t_{i,j}$ is $O(L M \Tmax)$: for every sequence
position and for every state in the HMM, we have to sum at most $\Tmax$ terms in order
to calculate the backward and forward terms.

(2) The $f(X_k, m)$ values are identical to the previously defined forward
probabilities and are calculated in the same way as in the forward algorithm.

(3) We now prove by induction that $t_{i,j}(X_k, l)$ is equal to the weighted
sum of probabilities of state paths that have at least one $i \rightarrow j$
transition and that finish at sequence position $k$ in state $l$, the weight
of each state path being equal to its number of $i \rightarrow j$ transitions.

Initialisation step (sequence position $k=0$): $t_{i,j}(X_0, m) = 0$ is true
as the sum of probabilities of state paths that finish in state $m$ at
sequence position $0$ and that have at least one $i \rightarrow j$ transition
is zero.  

Induction step $k \rightarrow k+1$: We now show that if Equation~\ref{eq:tij}
is true for sequence position $k$, it is also true for $k+1$. We have to
distinguish two cases:

(i) case $m = j$: 
\begin{eqnarray}
t_{i,j}(X_{k+1}, m) & = & f(X_k, i)T_{i,j} E_j(x_{k+1}) +  \label{eq:first} \\
                 &   & \sum_{n = 1}^M t_{i,j}(X_k, n) T_{n,j} E_j(X_{k+1}) \label{eq:second}
\end{eqnarray}

The first term, see right hand side of \ref{eq:first}, is the sum of
probabilities of state paths that finish at sequence position $k+1$ and whose
last transition is from $i \rightarrow j$. The second term, see
\ref{eq:second}, is the sum of probabilities of state paths that finish at
sequence position $k+1$ and that already have at least one $i \rightarrow j$
transition. Note that the term in \ref{eq:second} also contains a
contribution for $n = i$. This ensures that the weight of those state
path that already have at least one $i \rightarrow j$ transition is correctly
increased by 1.  The sum, $t_{i,j}(X_{k+1}, m)$, is therefore the weighted sum of
probabilities of state paths that finish in sequence position $k+1$ and 
contain at least one $i \rightarrow j$ transition. Each state path's weight
in the sum is equal to its number of $i \rightarrow j$ transitions.

(ii) case $m \ne j$:

\begin{equation}
t_{i,j}(X_{k+1}, m) = \sum_{n = 1}^M t_{i,j}(X_k, n) T_{n,m} E_{m}(x_{k+1}) \nonumber
\end{equation}

The expression on the right hand side is the weighted sum of probabilities of state
paths that finish in sequence position $k+1$ and contain at least one $i
\rightarrow j$ transition.

We have therefore shown that if Equation~\ref{eq:tij} is true for sequence
position $k$, it is also true for sequence position $k+1$. This concludes
the proof of theorem 1. $\Box$

\medskip

It is easy to show that $e_i(y, X)$ in Equation~\ref{eq:E_update} can also be
calculated in $O(M)$ memory and $O(L M \Tmax)$ time in a similar way as
$t_{i,j}(X)$.  Let $e_i(y, X_k, l)$ denote the weighted sum of probabilities
of state paths that finish at sequence position $k$ in state $l$ and for which
state $i$ reads letter $y$ at least once, the weight of each state path being
equal to the number of times state $i$ reads letter $y$. As in the calculation
of $t_{i,j}(X)$, we again omit the superfix $n$ as the calculation of $e_i(y,
X)$ is again entirely based on the transition and emission probabilities of
iteration $n$.

\textbf{Theorem 2:} $e_i(y, X)$ can be calculated in $O(M)$ memory and $O(L M
\Tmax)$ time using the following algorithm. $e_i(y, X)$ is the weighted sum of
probabilities of state paths for sequence $X$ that read letter $y$ in state $i$
at least once, the weight of each state path being equal to the number of
times letter $y$ is read by state $i$.

Initialisation step:

\begin{eqnarray}
f(X_0, m) &=& 
      \left\{ \begin{array}{ll} 
             1 & m = \textsl{Start} \nonumber \\
             0 & m \ne \textsl{Start} \nonumber
              \end{array}
      \right. \nonumber \\
e_i(y, X_0, m) &=& 0 \nonumber
\end{eqnarray}

Recursion:

\begin{eqnarray}
f(X_{k+1}, m) & = & \sum_{n = 1}^M f(X_k, n) T_{n,m} E_{m}(x_{k+1})\nonumber\\
e_i(y, X_{k+1}, m) & = & 
      \left\{ \begin{array}{ll} 
             \sum_{n = 1}^M e_i(y, X_k, n) T_{n,m} E_{m}(x_{k+1}) &  \\
                    \textrm{if} \quad m \ne i \quad \textrm{or} \quad x_{k+1} \ne y &  \\
                    & \\
             f(X_k, i) T_{i,m} E_m(x_{k+1}) +  & \\
             \sum_{n = 1}^M e_i(y, X_k, n) T_{n,m} E_m(x_{k+1}) & \\
                    \textrm{if} \quad m = i \quad \textrm{and} \quad x_{k+1} = y & 
              \end{array}
      \right. \nonumber
\end{eqnarray}

Termination step:

\begin{eqnarray}
P(X) = f(X_L, \textsl{End}) &=& \sum_{n = 1}^M f(X_L, n) T_{n,\textsl{End}} \label{eq:px2} \\
e_i(y, X) = e_i(y, X_L, \textsl{End}) & = & \sum_{n = 1}^M e_i(y, X_L, n) T_{n,\textsl{End}} \nonumber
\end{eqnarray}

\textbf{Proof:} The proof is strictly analogous to the proof of theorem 1.

%
%
%
%
%
%
%

\medskip

The above theorems have shown that $t_{i,j}(X)$ and $e_i(y, X)$ can each be
calculated in $O(M)$ memory and $O(L M \Tmax)$ time.  As there are $T$
transition parameters and $E$ emission parameters to be calculated in each
Baum-Welch iteration, and as these $T + E$ values can be calculated
independently, the time and memory requirements for each iteration and a set
of training sequences whose sum of sequence lengths is $L$ using our new
algorithm are

\begin{itemize}
\item $O(M)$ memory and $O(L M \Tmax\ (T + E))$ time, if all
      parameter estimates are calculated consecutively
\item $O(M (T + E))$ memory and $O(L M \Tmax)$ time, if all
      parameter estimates are calculated in parallel
\item more generally, $O(M c)$ memory and $O(L M \Tmax\ (T + E - c))$
      time for any $c \in \{1, \dots , (T + E)\}$, if $c$ of $T + E$
      parameters are to be calculated in parallel
\end{itemize}

Note that the calculation of $P(X)$ is a by-product of each $t_{i,j}(X)$ and
each $e_i(y, X)$ calculation, see Equations~\ref{eq:px} and \ref{eq:px2}, and
that $T$ is equal to the number of free transition parameters in the HMM which
is usually smaller than the number of transitions probabilities.  Likewise,
$E$ is the number of free emission parameters in the HMM which may differ from
the number of emission probabilities when the probabilities are parametrised.


\section*{Discussion and Conclusions}

We propose the first linear-memory algorithm for Baum-Welch training.  For a
hidden Markov model with $M$ states, $T$ free transition and $E$ free emission
parameters, and an input sequence of length $L$, our new algorithm requires
$O(M)$ memory and $O(L M \Tmax\ (T + E))$ time for one Baum-Welch
iteration as opposed to $O(\log(L) M)$ memory and $O(\log(L) L M \Tmax\ + L (T
+ E))$ time using the checkpointing algorithm \cite{ghs1997,th1998,
  wh2000}, where $\Tmax$ is the maximum number of states that any state is
connected to. Our algorithm can be generalised to pair-HMMs and, more
generally, n-HMMs that analyse n input sequences at a time in a
straightforward way.  In the n-HMM case, our algorithm reduces the memory and
time requirements from $O(\log(L) L^{n-1} M)$ memory and $O(\log(L) L^n M
\Tmax\ + L^n (T + E))$ time to $O(L^{n-1} M)$ memory and $O(L^n M \Tmax\ 
(T + E)))$ time. An added advantage of our new algorithm is that a reduced
time requirement can be traded for an increased memory requirement and
\emph{vice versa}, such that for any $c \in \{1, \dots, (T + E)\}$, a time
requirement of $L^n M \Tmax\ c$ incurs a memory requirement of $L^{n-1} M
(T + E - c)$. For HMMs, our novel algorithm renders the memory requirement
completely independent of the sequence length. Generally, for n-HMMs and all
$T + E$ parameters being estimated consecutively, our novel algorithm
reduces the memory requirement by a factor $\log(L)$ and the time requirement
by a factor $\log(L)/(T + E) + 1/(M \Tmax)$. For all hidden Markov models
whose number of states does not depend on the length of the input sequence,
this thus amounts to a significantly reduced memory requirement and --- in
cases where the number of free parameters and states of the model (i.e.\ $T +
E$) is smaller than the logarithm of sequence lengths --- even to a
reduced time requirement.

For example, for an HMM that is used to predict human genes, the training
sequences have a mean length of at least $2.7 \cdot 10^4$~bp which is the
average length of a human gene \cite{humangenome2001}. Using our new
algorithm, the memory requirement for Baum-Welch training is reduced by a
factor of about $28 \approx e * \ln{(2.7 \cdot 10^4)}$ with respect to the most
memory-sparse version of the checkpointing algorithm.

Our new algorithm makes use of the fact that the numerators and denominators
of Equations~\ref{eq:T_update} and \ref{eq:E_update} can be decomposed in a
smart way that allows a very memory-sparse calculation. This calculation
requires only one \emph{uni}-directional scan along the sequence rather than
one or more \emph{bi}-directional scans, see Figure~1. This property gives our
algorithm the added advantage that it is easier to implement as it does not
require programming techniques like recursive functions or checkpoints.

Baum-Welch training is only guaranteed to converge to a \emph{local} optimum.
Other optimisation techniques have been developed in order to find better
optima. One of the most successful methods is simulated annealing (SA)
\cite{kirkpatrick1983,dekm1998}. SA is essentially a Markov chain Monte Carlo
(MCMC) in which the target distribution is sequentially changed such that the
distribution gets eventually trapped in a local optimum. One can give proposal
steps a higher probability as they are approaching locally better points. This
can increase the performance of the MCMC method, especially in higher
dimensional spaces \cite{rr1998}. One could base the candidate distribution on
the expectations such that proposals are more likely to be made near the EM
updates (calculated with our algorithm). There is no need to update all the
parameters in one MCMC step, modifying a random subset of parameters yields
also an irreducible chain. The last feature makes SA significantly faster than
Baum-Welch updates as we need to calculate expectations only for a few
parameters using SA. In that way, our algorithm could be used for highly
efficient parameter training: using our algorithm to calculate the EM updates
in only linear space and using SA instead of the Baum-Welch algorithm for fast
parameter space exploration.

Typical biological sequence analyses these days often involve complicated
hidden Markov models such as pair-HMMs or long input sequences and we hope
that our novel algorithm will make Baum-Welch parameter training an appealing
and practicable option.

Other commonly employed methods in computer science and Bioinformatics are
stochastic context free grammars (SCFGs) which need $O(L^2 M)$ memory to
analyse an input sequence of length $L$ with a grammar having $M$ non-terminal
symbols \cite{dekm1998}. For a special type of SCFGs, known as covariance
models in Bioinformatics, $M$ is comparable to $L$, hence the memory
requirement is $O(L^3)$.  This has recently been reduced to $O(L^2 \log(L))$
using a divide-and-conquer technique \cite{eddy2002}, which is the SCFG
analogue of the Hirschberg algorithm for HMMs \cite{hirshberg1975}.  However,
as the states of SCFGs can generally impose long-range correlations between
any pair of sequence positions, it seems that our algorithm cannot be applied
to SCFGs in the general case.

\section*{Authors contributions}

The algorithm is the result of a brainstorming session of the authors on the
Genome campus bus back to Cambridge city centre on the evening of the 17th
February 2005. Both authors contributed equally.

\section*{Acknowledgements}

\ifthenelse{\boolean{publ}}{\small}{} 

The authors would like to thank one referee for the excellent comments.
I.M.\ is supported by a B\'ek\'esy Gy\"orgy postdoctoral fellowship.  Both
authors wish to thank Nick Goldman for inviting I.M.\ to Cambridge.

  
{\ifthenelse{\boolean{publ}}{\footnotesize}{\small}
\bibliographystyle{bmc_article}  
}

\ifthenelse{\boolean{publ}}{\end{multicols}}{}

\newpage

\section*{Figure}

\medskip

\subsection*{Figure 1 - Pictorial description of the new algorithm for pair-HMMs}

This figure shows a pictorial description of the differences between the
forward-backward algorithm (a) and our new algorithm (b) for the Baum-Welch
training of a pair-HMM. Each large rectangle corresponds to the projection of
the three-dimensional dynamic programming matrix (spanned by the two input
sequences $X$ and $Y$ and the states of the HMM) onto the sequence plane. (a)
shows how the numerator in Equation~\ref{eq:T_update} is calculated at the
pair of sequence positions indicated by the black square using the standard
forward and backward algorithm. (b) shows how our algorithm simultaneously
calculates a strip of forward values $f(X_k, Y_q, m)$ and a strip of
$t_{i,j}(X_k Y_q, m)$ values at sequence position $k$ in sequence $X$ in order
to estimate $t_{i,j}$ in Equation~\ref{eq:T_update}.

\bigskip

\begin{figure}[h]
\begin{center}
\epsfig{file=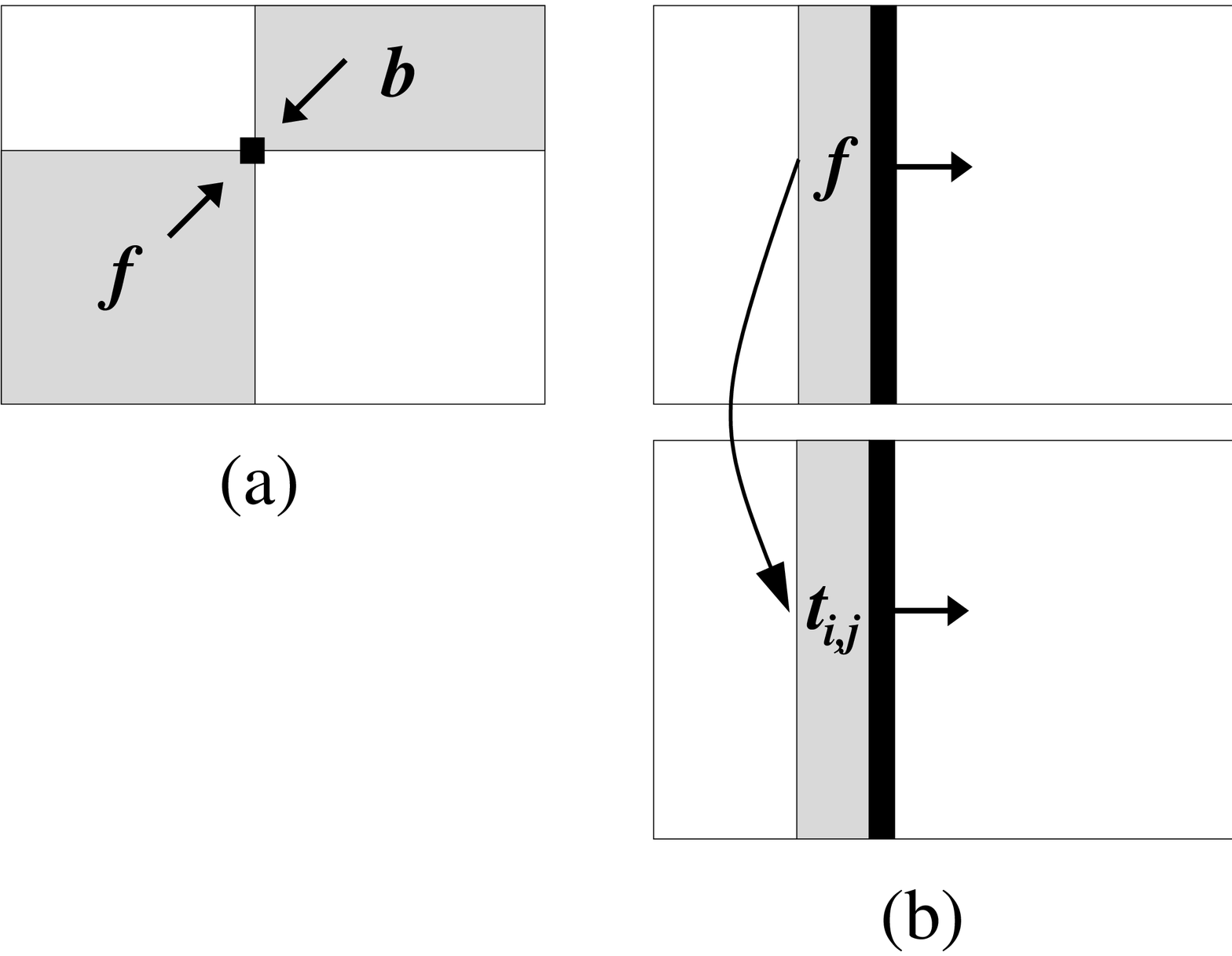,width=14.0cm}
\caption[]{
}
\end{center}
\end{figure}
 
\end{bmcformat}
\end{document}